\documentclass[10pt, a4paper]{article}
\usepackage{booktabs}
\usepackage{multirow}
\usepackage{multicol}
\usepackage{microtype}

\usepackage[final]{lrec2026} 

\title{Transfer Learning for Named Entity Recognition of Classical Latin through LLM Prompting}

\name{Callum Chan} 

\address{University of Ottawa \\
         Ottawa, Canada \\
         cchan073@uottawa.ca\\}

\abstract{With the increase in digitized resources of Classical Latin texts and modern breakthroughs of Large Language Models (LLMs), I contribute to ancient language research by participating in EvaLatin 2026. This paper describes Team uOttawa's system description and results for the Named Entity Recognition (NER) shared task. The task is divided into two subtasks: coarse-grained NER with 11 classes and fine-grained NER with 28 classes, each evaluated under strict and fuzzy regimes. Through prompt engineering of commercial LLMs gemini-2.5-pro and claude-sonnet-4-5, I show that the underrepresented ancient Latin language can take advantage of cross-lingual transfer learning by using advancements made by the wider LLM development community. Overall, the methods discussed in this report demonstrate very strong results, placing first in both NER subtasks and achieving the best scores across all evaluation metrics and regimes among all submissions.
 \\ \newline \Keywords{Named Entity Recognition, Statistical and Machine Learning Methods} }

\begin{document}

\maketitleabstract


\section{Introduction}

The study of computational linguistics for ancient languages is hampered from the lack of large annotated datasets \citep{sprugnoli-etal-2020-overview}. However, Ancient Natural Language Processing (ANLP) researchers have identified that modern state-of-the-art general purpose Large Language Models (LLMs) trained on vast amounts of multilingual data have great potential to bridge this low-resource gap \citep{zhong2026opportunitieschallengeslargelanguage}. The ANLP community has contributed resources for Classical Latin NER \cite{romanello2024named}, alongside work applying modern deep learning methods to automate the task \cite{erdmann-etal-2019-practical}. ANLP practitioners have also leveraged contemporary BERT \cite{devlin2019bertpretrainingdeepbidirectional} and embedding-based NLP techniques, yielding promising results. Nevertheless, automated NER of Classical Latin texts remains an open research problem \cite{bamman2020latinbertcontextuallanguage,riemenschneider-frank-2023-exploring,torres-aguilar-2022-multilingual}.

Through transfer learning, generative auto-regressive LLMs have shown strong capabilities in natural language understanding and generation across languages for which there are limited resources \cite{zhong2026opportunitieschallengeslargelanguage}. It has also been discussed that tasks involving ancient Latin texts benefit from these advancements \cite{volk-etal-2024-llm}. In particular, prompting ChatGPT has recently been examined as a potential approach to NER for ancient languages \cite{woodrum2025named}.

These studies, as well as my own experiences, lead me to explore prompting strategies on generative LLMs for this shared task. Examining literature on LLM prompting, it is widely understood that few-shot techniques help guide these models to understand given assignments by providing context and examples \cite{brown2020languagemodelsfewshotlearners}. In particular, thoughtful examples lead to improved results in terms of both output format and content across a wide range of tasks through in-context learning \cite{liu2021pretrainpromptpredictsystematic}.

By participating in this shared task as Team uOttawa, I present the ANLP community with another avenue for automated NER of ancient texts. My submissions show that commercial LLMs have the potential to accurately identify named entities of Latin text through principled prompting strategies. Team uOttawa places first for all subtasks and achieves the best scores across all evaluation methods for EvaLatin's 2026 NER shared task.

\section{Shared Task}
The EvaLatin 2026 NER shared task consists of two subtasks and two evaluation regimes. Each subtask charged participants to identify and classify named entities  within a corpus of Classical Latin text. The first subtask focused on coarse-grained named entity recognition; a multi-class classification task consisting of 11 broad classes. The second subtask increased the granularity of the classification by introducing subcategories of the 11 classes mentioned previously. This fine-grained multi-class classification consisted of 28 labels. These classes are shown in Table ~\ref{tab:classes}. Per the task guidelines, all text is tokenized and each token is given a named entity label in the Inside-Outside-Beginning (IOB) format.

\begin{table}[ht]
    \centering
    \scalebox{0.8}{
    \begin{tabular}{l|ll}
        \textbf{Coarse} & \multicolumn{2}{c}{\textbf{Fine}}  \\
        \hline
        Person & Person & Collective.ancestry \\
        Place & Person.author & Collective.organization \\
        Collective & Person.ancestry & Collective.epithet \\
        Creature & Person.epithet & Collective.derivative \\
        Event & Person.ethnic & Creature \\
        Language & Person.derivative & Creature.animal \\
        NCE & Place & Creature.astronomy \\
        Object & Place.astronomy & Event \\
        Misc & Place.epithet & Language \\
        Time & Place.derivative & NCE \\
        Work & Collective & Object \\
        & Collective.animal & Misc\\
        & Collective.astronomy & Time \\
        & Collective.ethnic & Work \\
    \end{tabular}}
    \caption{Classes of coarse and fine-grained classification subtasks for EvaLatin 2026 Named Entity Recogintion. NCE stands for Non-Consecutive-Entity. }
    \label{tab:classes}
\end{table}

\section{Evaluation Metrics}
Each subtask was evaluated under a strict and fuzzy evaluation regime. Strict evaluation considered exact matching of entity types and token boundaries, whereas the fuzzy evaluation loosened token boundaries and allowed for mis-identification of named entities so long as one token overlapped with the gold standard. With respect to metrics, entity level, micro precision, recall and F1 measure were used to score submissions.

\section{Dataset}
\label{dataset}
A dataset consisting of around 2900 annotated tokens was provided by the shared task organizers as sample data. This text was taken from poetic and prose textual sources and contained 86 named entities belonging to the `Collective', `Misc', `Person' and `Place' classes.

In addition to the data provided by the organizers, I also collected a corpus of annotated named Latin entities from the works of \cite{torres-aguilar-2022-multilingual} and \cite{erdmann-etal-2019-practical}.
In all, this supplemental dataset consisted of 1.48M classified Latin tokens. Due to differences in class naming, this text was pre-processed to align with the tagset used by the shared task. The mapping between supplementary and EvaLatin 2026 tags is illustrated in Table \ref{tab:tagset}. However, the supplemental dataset suffers from limited tagset coverage, containing only 3 of the 11 coarse-grained tags and lacks more detailed, fine-grained classes. This dataset was used both in examples for my prompting strategies from Section \ref{description} and in additional experiments described in Appendix \ref{appendix_extra_experiments}.

\begin{table}[ht]
    \centering
    \scalebox{0.8}{
    \begin{tabular}{l|l}
        \textbf{Supplementary Tag} & \textbf{EvaLatin 2026 Tag}  \\
        \hline
        PERS, PRS & Person \\
        LOC, GEO & Place \\
        GRP & Collective \\
    \end{tabular}}
    \caption{Tagset alignment between supplementary dataset and EvaLating 2026 NER tagset. }
    \label{tab:tagset}
\end{table}

The test dataset for official scoring consisted of around 26,000 tokens which were sampled from works from Tacitus, Pliny the Elder and Ovid. This selection was, again, chosen to give systems a diverse array of poetic and prose texts. This test set also contained a wider array of named entities, including more complex references to geographical, demographical, ethnical, mythological and humanities-related entities when compared to the sample data.

\section{Description of the system}
\label{description}
To tackle each of the subtasks of the EvaLatin 2026 NER shared task, I employed prompting techniques on public instances of gemini-2.5-pro and claude-sonnet-4-5. These models were chosen for their predecessor's strong generative performance and multilingual understanding \cite{ahuja2024megaverse,jayakody2024performance}.
For each subtask (coarse and fine) two distinct prompting strategies were used: zero-shot learning and few-shot learning. Prompt templates were created for each strategy and each strategy was run against both subtasks. In all, 4 prompt templates were created to perform experiments to address this shared task. Each of the prompt templates included the following sections: 
\begin{itemize}
    \item Task overview, which consisted of the broad task guidelines.
    \item Task specifics, which consisted of the specific task requirements.
    \item Types of entities, which listed and described each of the named entity classes.
    \item Guidelines for output, which specified the desired format of responses.
\end{itemize}

The primary difference between the subtask prompts lies in the scope of their entity descriptions, with the coarse-grained prompts covering 11 classes and the fine-grained prompts providing explanations for all 28.

With respect to prompting strategies, I employed zero-shot and few-shot techniques. Both techniques contained the same sections, however few-shot prompts also included 3 example sentences of Latin NER in the desired output format. These examples were randomly sampled from the larger supplementary corpus.

The input text was split into sentences and each sentence was given its own prompt. Then, each prompt was fed into the models under consideration and responses were collected. Some light post-processing of responses was conducted to clean up the formatting and, if a token was not given a label by the model, it was automatically classified as not belonging to a named entity. All 4 prompt templates used in these methods are listed in Appendix \ref{appendix_prompts}.

\subsection{Additional Experiments}

Additional experiments were performed after the official scoring period of EvaLatin 2026. These methods explore fine-tuning Latin BERT-based models for token classification. They were not submitted for official scoring due to time-constraints of the evaluation period and the limitation of 2 submissions per team. Their details and results are briefly described in Appendix \ref{appendix_extra_experiments}.

\section{Results}
The results of my experimentation and official submissions show that LLMs and thoughtful prompting can be effectively applied to the task of textual Latin NER. Before the official scoring period of EvaLatin 2026, I performed preliminary assessments of these methods on the sample data provided by the shared task organizers. These results are shown in in Table ~\ref{tab:prelimresults} and were scored using the official HIPE 2020 scorer\footnote{\url{https://github.com/hipe-eval/HIPE-scorer}}.

Overall, comparing prompt strategies across models, few-shot techniques consistently outperform zero-shot prompts. Furthermore, gemini-2.5-pro generally performs better at fine-grained classification, whereas claude-sonnet-4-5 outperforms in the coarse-grained classification subtask.

Using these observations, I submitted official predictions on the test set using the few-shot prompting strategy for both models, across both subtasks, for evaluation. Submission uOttawa\_nerc\_1 contains the responses from gemini-2.5-pro and uOttawa\_nerc\_2 contains the responses from claude-sonnet-4-5. All official scoring was performed by the shared task organizers and the results of my submissions are reported in Table ~\ref{tab:officialresults}. Consistent with preliminary experimental results, gemini-2.5-pro demonstrates superior performance in fine-grained classification, whereas claude-sonnet-4-5 excels at the coarse-grained level.

Both uOttawa\_nerc\_1 and uOttawa\_nerc\_2 outperformed all other submissions across precision, recall and F1 measure. This disparity is also presented in Table ~\ref{tab:officialresults} where next best scores by submissions KULeuven\_nerc-coarse\_1 and argo-navis\_nerc-fine\_1 are reported. While the gap in performance is modest, around 0.1 to 0.2 for the coarse-grained subtask, it is much more pronounced for the fine-grained subtask with recall and F1 measures lagging by up to 0.5.

These findings demonstrate that careful prompting of commercial LLMs can provide ANLP researchers with powerful ways to automate NER for Classical Latin text.

\begin{table*}[h!]
    \centering
    \begin{tabular}{|c|c||ccc||ccc|}
        \cline{3-8}
         \multicolumn{2}{c||}{ } & \multicolumn{3}{c||}{\textbf{Coarse}} & \multicolumn{3}{c|}{\textbf{Fine}} \\
        \hline
        \textbf{Regime} & \textbf{Method} & \textbf{P} & \textbf{R} & \textbf{F1} & \textbf{P} & \textbf{R} & \textbf{F1} \\
        \hline
        \multirow{4}{*}{Strict}
            & gemini zero-shot & 0.716 & \textbf{0.839} & 0.772 & 0.619 & \textbf{0.805} & \underline{0.700} \\
            & gemini few-shot & 0.737 & \textbf{0.839} & 0.785 & 0.650 & \underline{0.770} & \textbf{0.705} \\
            & sonnet zero-shot & \underline{0.783} & 0.828 & \underline{0.804} & \underline{0.667} & 0.621 & 0.643 \\
            & sonnet few-shot & \textbf{0.849} & \textbf{0.839} & \textbf{0.844} & \textbf{0.698} & 0.690 & 0.694 \\
        \hline
        \multirow{4}{*}{Fuzzy}
            & gemini zero-shot & 0.755 & \underline{0.885} & 0.815 & 0.628 & \textbf{0.816} & 0.710 \\
            & gemini few-shot & 0.788 & \textbf{0.897} & \underline{0.839} & \underline{0.689} & \textbf{0.816} & \underline{0.747} \\
            & sonnet zero-shot & \underline{0.815} & 0.862 & 0.838 & 0.679 & 0.632 & 0.655 \\
            & sonnet few-shot & \textbf{0.895} & \underline{0.885} & \textbf{0.890} & \textbf{0.756} & 0.747 & \textbf{0.751} \\
        \hline
    \end{tabular}
    \caption{Preliminary results of Team uOttawa's methods on the sample annotated dataset provided by the shared task organizers. Top results have been \textbf{bolded}, and the next-best scores have been \underline{underlined}. }
    \label{tab:prelimresults}
\end{table*}

\begin{table*}[h!]
    \centering
    \begin{tabular}{|c|c||ccc||ccc|}
        \cline{3-8}
         \multicolumn{2}{c||}{ } & \multicolumn{3}{c||}{\textbf{Coarse}} & \multicolumn{3}{c|}{\textbf{Fine}} \\
        \hline
        \textbf{Regime} & \textbf{Submission} & \textbf{P} & \textbf{R} & \textbf{F1} & \textbf{P} & \textbf{R} & \textbf{F1} \\
        \hline
        \multirow{2}{*}{Strict}
            & uOttawa\_nerc\_1
            &  0.876 & 0.914 & 0.895 & \textbf{0.841} & \textbf{0.890} & \textbf{0.865} \\
            & uOttawa\_nerc\_2
            & \textbf{0.899} & \textbf{0.917} & \textbf{0.908} & \textbf{0.841} & 0.871 & 0.856 \\
            &\textit{next\_best\_strict} & \textit{0.736} & \textit{0.694} & \textit{0.714} & \textit{0.466} & \textit{0.327} & \textit{0.384} \\
        \hline
        \multirow{2}{*}{Fuzzy}
            & uOttawa\_nerc\_1 
            & 0.917 & \textbf{0.956} & 0.936 & 0.862 & \textbf{0.913} & \textbf{0.887} \\
            & uOttawa\_nerc\_2 
            & \textbf{0.932} & 0.950 & \textbf{0.941} & \textbf{0.863} & 0.893 & 0.878 \\
            & \textit{next\_best\_fuzzy} & \textit{0.794} & \textit{0.749} & \textit{0.771} & \textit{0.523} & \textit{0.367} & \textit{0.432} \\
        \hline
    \end{tabular}
    \caption{Official results for Team uOttawa's submissions. Top results have been \textbf{bolded}. Each submission contained predictions for both coarse and fine-grained subtasks. The uOttawa\_nerc\_1 submission employed few-shot prompting on gemini-2.5-pro and the uOttawa\_nerc\_2 submission employed few-shot prompting on claude-sonnet-4-5. The next best scores from other teams' submission to the shared task are also reported as next\_best\_strict and next\_best\_fuzzy from submissions KULeuven\_nerc-coarse\_1 and argo-navis\_nerc-fine\_1.}
    \label{tab:officialresults}
\end{table*}

\section{Discussion}
Overall, we can see that the results of Team uOttawa's methods across subtasks and evaluation regimes is competitive. 

With respect to intra-model performance over the sample data, I note that few-shot techniques outperformed its zero-shot counterparts in nearly every single metric, across both subtasks and for all evaluation regimes. This is consistent across both models. It is shown that the performance gained from the technique of few-shot learning is model agnostic and clearly benefits both LLMs for the task of Latin NER. Zero-shot prompting does not provide enough context to meaningfully activate the models' latent understanding of Latin text when compared to few-shot techniques. The presence of examples that include output format, Latin text, and their annotated named entities guides the model in its understanding of the task. The increase in contextual depth to the prompt (i.e. the starting point) is crucial in the iterative generation loop performed by modern auto-regressive LLMs. Since Classical Latin is a low-resource language, when compared to English, the explicit guidance given by examples helps to alleviate the gap in understanding brought about by the models' weaker internal representation of Latin grammar, morphology and named entity conventions. Examples provide the model with domain cues to help in its classification with respect to understanding what language it should use to perform this task.

Another observation warranting further discussion are the high recall scores. I notice that across both sample and test datasets, coarse and fine-grained classification, gemini and sonnet models, prompting techniques and evaluation regimes, recall scores are consistently higher than precision. These results illustrate my method's capacity to identify named entities with few false negatives. However, this also demonstrates a tendency toward over-generation or over-classification. By disproportionately labeling Latin tokens as named entities, my method minimizes missed detections at the cost of incorrectly flagging non-named entities as named entities. This phenomenon stems from the tendency of English and other modern languages to adopt Latin common nouns as proper names. A primary example is the word `Gloria'; while modern English categorizes this exclusively as a name, in Latin it functions strictly as a common noun meaning `glory.' Because modern commercial LLMs are trained on vast English and Romance-language corpora, their named entity recognition is inevitably biased toward these derivative languages, leading to the misapplication of modern entity structures over those native to Latin.

Finally, I notice that scores for the fine-grained classification task are poorer across both models, prompting techniques and evaluation regimes when compared to the coarse-grained subtask. This suggests that while few-shot prompting yields measurable improvements, highly nuanced classification tasks involving many labels remain a significant challenge for LLMs. This is an indication that these highly generalized commercial language models are able to classify broad categories of named entities, which may have been present in their training data. However, their understanding of natural language does not translate as well when presented with new categories or unseen finer-grained labels.

In summary, I have discussed that my methods demonstrate the effectiveness of prompt engineering with commercial LLMs for the task of Latin NER. From my results, it has been shown that few-shot techniques guide these models to accurately and precisely identify and classify named entities in Latin. 

\section{Practical Adoption}

The methods described in this paper presents powerful tools to the greater ANLP community. They both achieve impressive results and are widely accessible to those unfamiliar with machine learning or computational linguistic technicalities. A pipeline consisting of few-shot LLM prompting over non-classified Latin text along with a second pass of humans-in-the-loop (HITL) could prove valuable to researchers studying these documents. My results also show that human correctors should concentrate their efforts on identifying tokens misidentified as named entities  (because of the high recall scores) and on instances of rare or unusual labels. 

\section{Future Work}
My work demonstrates that there exists powerful tools for automatically classified Latin NER. I also recognize that this task is far from solved and there is room to improve these results. Latin is morphologically complex and textual fine-grained multi-label classification over its tokens remains challenging.  

My main methods focus on prompting auto-regressive commercial LLMs for token classification. Future work could include exploring LLM hyperparmeter tuning of temperature or top\_p. Probing various other open source models and their fine-tuning over Latin text could also lead to improved results. With respect to the prompting techniques, a wider investigation of prompting strategies should be performed. In particular, I express that there could be more intentional ways of choosing examples that are given within prompts. A cosine similarity-based approach as described in \cite{liu-etal-2022-makes} could benefit classification performance by identifying the most relevant examples for a given input sentence, more so than a random selection as was performed in my methods.

Finally, there has been progress in training BERT based models over Latin text \cite{bamman2020latinbertcontextuallanguage,riemenschneider-frank-2023-exploring,torres-aguilar-2022-multilingual}. 
It has also been shown that the fine-tuning of these architectures still perform competitively for ancient language classification tasks \cite{sommerschield-etal-2023-machine}.
Given the increase in digitized annotated Latin resources, more concentrated efforts than my additional experiments to fine-tune these embedding-based models could yield better results at the fraction of the cost it would take to train or fine-tune a generative LLM.  

\section{Conclusion}
To conclude, I have showcased that the underrepresented ancient Latin language benefits from transfer learning of commercial generative LLMs gemini-2.5-pro and claude-sonnet-4-5. Through few-shot prompting of these modern models, Team uOttawa achieves strong performance when compared to all submissions made to the EvaLatin 2026 NER shared task.
I encourage ANLP researchers to continue exploring advancements made by the wider computational linguistic community that bridge the low-resource gap for tasks involving Classical Latin.

\section{Limitations}
My results are limited to the text provided by the shared task organizers. Although the text has been chosen to vary with respect to genre and author, it is far from representative of the greater body of works that comprise all known Latin corpora.

Furthermore, I recognize that my experiments have been performed on a limited number of LLMs, gemini-2.5-pro and claude-sonnet-4-5. Ablation studies on prompt variations or hyperparameter selection and tests for statistical significance were not performed due to the limited time frame of this shared task. 

Finally, my prompt formulations were influenced by my prior experience working with LLMs across different domains and for different purposes. These experiences could introduce biases into the prompting strategy and skew the results of my experiments.

\section{Bibliographical References}\label{sec:reference}

\bibliographystyle{lrec2026-natbib}
\bibliography{lrec2026-NER}

\label{lr:ref}
\bibliographystylelanguageresource{lrec2026-natbib}

\appendix
\section{Appendix}
\label{appendix_label}

\subsection{Additional Experiments}
\label{appendix_extra_experiments}
A brief exploration of fine-tuning BERT-based models for token classification was performed. Building on the works of \cite{riemenschneider-frank-2023-exploring}, I fine-tuned their LaBERTa model on the supplemental dataset described in Section \ref{dataset}. This dataset contained 15,593 sentences and an 80-20 train-validation split was performed. A learning rate of 2e-5 was used with a weight decay of 0.01 over 10 epochs. From these epochs the best model was saved for inference. This training was conducted using HuggingFace's Trainer framework \footnote{\url{https://huggingface.co/docs/transformers/en/main_classes/trainer}}.

Since the tags of the supplemental dataset only contains `Person', `Place' and `Collective' classes, this method was only used to generate predictions for the coarse-grained classification subtask.

\begin{table}[h!]
\centering
\begin{tabular}{|c|c|c|c|}
\hline
\textbf{Regime} & \textbf{P} & \textbf{R} & \textbf{F1}\\
\hline
Strict & 0.735 & 0.763 & 0.749 \\
Fuzzy & 0.770 & 0.800 & 0.785 \\
\hline
\end{tabular}
\caption{Results of the extra experiments performed by fine-tuning laBERTa on the supplemental dataset. Scores are from the coarse-grained classification subtask.}
\label{tab:extraresults}
\end{table}

While not as strong as the results for my main methods, these scores are still quite competitive. With respect to F1 measure, this submission would have placed third behind uOttawa\_nerc\_2 and uOttawa\_nerc\_1. In this instance, prompting large commercial LLMs outperforms embedding-based classification.

\onecolumn
\subsection{Prompt Templates}
\label{appendix_prompts}
The following appendix section presents the four prompt templates that were used for my submissions to the EvaLatin 2026 NER shared task. These are shown in Tables \ref{zero-coarse}, \ref{few-coarse}, \ref{zero-fine} and \ref{few-fine}.

\begin{table}[h!]
\centering
\scalebox{0.6}{\begin{tabular}{|c|}
\hline
\multicolumn{1}{p{24cm}}{
\#\#\# Task Outline \newline
Your task is to perform coarse grained named entity recognition on latin text. The latin text is already tokenized and your job is to identify tokens that belong to named entities. \newline
 \newline
--- \newline
 \newline
\#\#\# Task Specifics \newline
- Named entity recognition is a technique that identifies and classifies entities in written language into categories such as names, organizations, locations, time expressions, quantites and more. \newline
- You are to use the IOB (Inside Outside Beginning) format to clasify each token as either B - the first token of a multi-token entity, I - any subsequent token within the same entity, or O - a token that does not belong to an entity. \newline
- If the token is idenified as I or B, you are then tasked with identifying the type of entity it is. I or B should be followed by a dash '-' then by the type label. \newline
- The input text will contain each token of the sentence on a newline. \newline
 \newline
--- \newline
 \newline
\#\#\# Types of Entities (Labels) \newline
The following is the list of the entity labels and their definition: \newline
- Person: Any identifiable single individual, including deities and anthropomorphic mythological figures. \newline
- Place: A politically, culturally, or geographically defined location, including fictional spaces and structures like temples, buildings, specific urban areas (e.g., gymnasia), and houses. \newline
- Collective: A named group of people or other creatures with shared identifiable characteristics on social, intellectual, political, national, family, mythical, or ethnic basis. \newline
- Creature: Mythical or real precisely identifiable non-human, non-anthropomorphic creatures. \newline
- Event: Significant named events identified by a string with a precise boundary. \newline
- Language: Languages and dialects clearly identified as such. \newline
- Non-consecutive-entity: Non-consecutive entities are strings that contain one named entity but are split across the text. These are provisionally tagged as non-consecutive-entity. \newline
- Object: Artifacts or groups of artifacts clearly identified with a name, such as ships, weapons, statues, columns, dedications, etc. \newline
- Miscellaneous: Entities that do not (yet) have a specific first-level tag among those provided. \newline
- Time: Any absolute date or time expression. \newline
- Work: Titles of literary or non-literary works, in any form. \newline
 \newline
--- \newline
 \newline
\#\#\# Guidelines for Output: \newline
- For each word in the sentence, output the word followed by its IOB label and entity type where appropriate. \newline
- Use only the set of labels as defined earlier. \newline
- Each word should be on a new line in the same way it was input. \newline
- Do not add any other explanations or text. \newline
 \newline
--- \newline
 \newline
\#\#\# Input: \newline
\{INPUT\_SENTENCE\} \newline
 \newline
\#\#\# Output: \newline
} \\
\hline
\end{tabular}}
\caption{Zero-Shot prompt template for coarse-grained NER.}
\label{zero-coarse}
\end{table}

\begin{table}[h!]
\centering
\scalebox{0.6}{\begin{tabular}{|c|}
\hline
\multicolumn{1}{p{24cm}}{
\#\#\# Task Outline \newline
Your task is to perform coarse grained named entity recognition on latin text. The latin text is already tokenized and your job is to identify tokens that belong to named entities.\newline
\newline
---\newline
\newline
\#\#\# Task Specifics\newline
- Named entity recognition is a technique that identifies and classifies entities in written language into categories such as names, organizations, locations, time expressions, quantites and more.\newline
- You are to use the IOB (Inside Outside Beginning) format to clasify each token as either B - the first token of a multi-token entity, I - any subsequent token within the same entity, or O - a token that does not belong to an entity.\newline
- If the token is idenified as I or B, you are then tasked with identifying the type of entity it is. I or B should be followed by a dash '-' then by the type label.\newline
- The input text will contain each token of the sentence on a newline.\newline
\newline
---\newline
\newline
\#\#\# Types of Entities (Labels)\newline
The following is the list of the entity labels and their definition:\newline
- Person: Any identifiable single individual, including deities and anthropomorphic mythological figures.\newline
- Place: A politically, culturally, or geographically defined location, including fictional spaces and structures like temples, buildings, specific urban areas (e.g., gymnasia), and houses.\newline
- Collective: A named group of people or other creatures with shared identifiable characteristics on social, intellectual, political, national, family, mythical, or ethnic basis.\newline
- Creature: Mythical or real precisely identifiable non-human, non-anthropomorphic creatures.\newline
- Event: Significant named events identified by a string with a precise boundary.
- Language: Languages and dialects clearly identified as such.\newline
- Non-consecutive-entity: Non-consecutive entities are strings that contain one named entity but are split across the text. These are provisionally tagged as non-consecutive-entity.\newline
- Object: Artifacts or groups of artifacts clearly identified with a name, such as ships, weapons, statues, columns, dedications, etc.\newline
- Miscellaneous: Entities that do not (yet) have a specific first-level tag among those provided.\newline
- Time: Any absolute date or time expression.\newline
- Work: Titles of literary or non-literary works, in any form.\newline
\newline
---\newline
\newline
\#\#\# Guidelines for Output:\newline
- For each word in the sentence, output the word followed by its IOB label and entity type where appropriate.\newline
- Use only the set of labels as defined earlier.\newline
- Each word should be on a new line in the same way it was input.\newline
- Do not add any other explanations or text.\newline
\newline
---\newline
\newline
\#\#\# Example:\newline
\newline
-> Example 1\newline
\#\#\#\#\# Input:\newline
\{INPUT\_EXAMPLE\_1\}\newline
\newline
\#\#\#\#\# Output:\newline
\{OUTPUT\_EXAMPLE\_1\}\newline
\newline
-> Example 2\newline
\#\#\#\#\# Input:\newline
\{INPUT\_EXAMPLE\_2\}\newline
\newline
\#\#\#\#\# Output:\newline
\{OUTPUT\_EXAMPLE\_2\}\newline
\newline
-> Example 3\newline
\#\#\#\#\# Input:\newline
\{INPUT\_EXAMPLE\_3\}\newline
\newline
\#\#\#\#\# Output:\newline
\{OUTPUT\_EXAMPLE\_3\}\newline
\newline
---\newline
\newline
\#\#\# Input:\newline
\{INPUT\_SENTENCE\}\newline
\newline
\#\#\# Output:\newline
} \\
\hline
\end{tabular}}
\caption{Few-Shot prompt template for coarse-grained NER.}
\label{few-coarse}
\end{table}

\begin{table}[h!]
\centering
\scalebox{0.6}{\begin{tabular}{|c|}
\hline
\multicolumn{1}{p{24cm}}{
\#\#\# Task Outline \newline
Your task is to perform fine grained named entity recognition on latin text. The latin text is already tokenized and your job is to identify tokens that belong to named entities.\newline
\newline
---\newline
\newline
\#\#\# Task Specifics\newline
- Named entity recognition is a technique that identifies and classifies entities in written language into categories such as names, organizations, locations, time expressions, quantites and more.\newline
- You are to use the IOB (Inside Outside Beginning) format to clasify each token as either B - the first token of a multi-token entity, I - any subsequent token within the same entity, or O - a token that does not belong to an entity.\newline
- If the token is idenified as I or B, you are then tasked with identifying the first level type of entity it is. I or B should be followed by a dash '-' then by the first level tag.\newline
- Furthermore, if the token belongs to a set of second level tags (defined below), the first level tag should be followed by a pont "." then by the second level tag.\newline
- The input text will contain each token of the sentence on a newline.\newline
\newline
---\newline
\newline
\#\#\# First Level Tags\newline
The following is the list of the first level entity labels, definitions and available second level tags:\newline
- Person: Any identifiable single individual, including deities and anthropomorphic mythological figures.\newline
-- .author\newline
-- .ancestry\newline
-- .epithet\newline
-- .ethnic\newline
-- .derivative\newline
- Place: A politically, culturally, or geographically defined location, including fictional spaces and structures like temples, buildings, specific urban areas (e.g., gymnasia), and houses.\newline
-- .astronomy\newline
-- .epithet\newline
-- .derivative\newline
- Collective: A named group of people or other creatures with shared identifiable characteristics on social, intellectual, political, national, family, mythical, or ethnic basis.\newline
-- .ancestry\newline
-- .animal\newline
-- .astronomy\newline
-- .ethnic\newline
-- .organization\newline
-- .epithet\newline
-- .derivative\newline
- Creature: Mythical or real precisely identifiable non-human, non-anthropomorphic creatures.\newline
-- .animal\newline
-- .astronomy\newline
- Event: Significant named events identified by a string with a precise boundary.\newline
- Language: Languages and dialects clearly identified as such.\newline
- Non-consecutive-entity: Non-consecutive entities are strings that contain one named entity but are split across the text. These are provisionally tagged as non-consecutive-entity.\newline
- Object: Artifacts or groups of artifacts clearly identified with a name, such as ships, weapons, statues, columns, dedications, etc.\newline
- Miscellaneous: Entities that do not (yet) have a specific first-level tag among those provided.\newline
- Time: Any absolute date or time expression.\newline
- Work: Titles of literary or non-literary works, in any form.\newline
\newline
\#\#\# Second Level Tags\newline
The following is the list of second level labels and their description:\newline
- .ancestry (collective.ancestry, person.ancestry): A designation or expression that refers unambiguously to one individual or group of individuals by using a family name, patronymic, matronymic, or other indication of lineage or familial relationships.\newline
- .animal (collective.animal, creature.animal): A type of creature or collective of creatures clearly identifiable with an animal or animal species.\newline
- .astronomy (creature.astronomy, collective.astronomy, place.astronomy): Named stars, groups of stars, constellations, and planets.\newline
- .author (person.author): A person clearly mentioned in relation to works they have authored. This tag may be modified or even omitted for project-specific goals.\newline
- .derivative (collective.derivative, person.derivative, place.derivative): An adjective derived from a toponym, personal name, or group name, used to identify things that are not individuals or collectives (for individuals or collectives, see .ethnic). Only the derivative is annotated, as the common noun in the expression does not act as a rigid designator. The first-level tag depends on the name from which the adjective derives (e.g. “Iberian” will be a place, “Platonic” a person, etc.).\newline
- .ethnic (collective.ethnic, person.ethnic): An ethnonym, demonym, or other word used to identify persons or collectives by means of their membership to a geographically or ethnically defined group. This tag is exclusively used with persons or collectives, as ethnics are mainly used in the ancient world to identify individuals via ethnic memberships (see also our rationale below). For all other uses of adjectives derived from places, use the .derivative subtag.\newline
- .epithet (collective.epithet, person.epithet, place.epithet): A capitalized epithet used to refer unambiguously to one individual, location, or collective, including nicknames, titles, and other appellatives.\newline
- .organization (collective.organization): Collectives identified by precise organizational structures, such as priesthoods, legions, religious, intellectual, or political groups and institutions, and so on.\newline
\newline
---\newline
\newline
\#\#\# Guidelines for Output:\newline
- For each word in the sentence, output the word followed by its IOB label and entity type where appropriate.\newline
- Use only the set of labels as defined earlier.\newline
- Each word should be on a new line in the same way it was input.\newline
- Do not add any other explanations or text.\newline
\newline
---\newline
\newline
\#\#\# Input:\newline
\{INPUT\_SENTENCE\}\newline
\newline
\#\#\# Output:\newline
} \\
\hline
\end{tabular}}
\caption{Zero-Shot prompt template for fine-grained NER.}
\label{zero-fine}
\end{table}

\begin{table}[h!]
\centering
\scalebox{0.5}{\begin{tabular}{|c|}
\hline
\multicolumn{1}{p{24cm}}{
\#\#\# Task Outline \newline
Your task is to perform fine grained named entity recognition on latin text. The latin text is already tokenized and your job is to identify tokens that belong to named entities.\newline
\newline
---\newline
\newline
\#\#\# Task Specifics\newline
- Named entity recognition is a technique that identifies and classifies entities in written language into categories such as names, organizations, locations, time expressions, quantites and more.\newline
- You are to use the IOB (Inside Outside Beginning) format to clasify each token as either B - the first token of a multi-token entity, I - any subsequent token within the same entity, or O - a token that does not belong to an entity.\newline
- If the token is idenified as I or B, you are then tasked with identifying the first level type of entity it is. I or B should be followed by a dash '-' then by the first level tag.\newline
- Furthermore, if the token belongs to a set of second level tags (defined below), the first level tag should be followed by a pont "." then by the second level tag.\newline
- The input text will contain each token of the sentence on a newline.\newline
\newline
---\newline
\newline
\#\#\# First Level Tags\newline
The following is the list of the first level entity labels, definitions and available second level tags:\newline
- Person: Any identifiable single individual, including deities and anthropomorphic mythological figures.\newline
-- .author\newline
-- .ancestry\newline
-- .epithet\newline
-- .ethnic\newline
-- .derivative\newline
- Place: A politically, culturally, or geographically defined location, including fictional spaces and structures like temples, buildings, specific urban areas (e.g., gymnasia), and houses.\newline
-- .astronomy\newline
-- .epithet\newline
-- .derivative\newline
- Collective: A named group of people or other creatures with shared identifiable characteristics on social, intellectual, political, national, family, mythical, or ethnic basis.\newline
-- .ancestry\newline
-- .animal\newline
-- .astronomy\newline
-- .ethnic\newline
-- .organization\newline
-- .epithet\newline
-- .derivative\newline
- Creature: Mythical or real precisely identifiable non-human, non-anthropomorphic creatures.\newline
-- .animal\newline
-- .astronomy\newline
- Event: Significant named events identified by a string with a precise boundary.\newline
- Language: Languages and dialects clearly identified as such.\newline
- Non-consecutive-entity: Non-consecutive entities are strings that contain one named entity but are split across the text. These are provisionally tagged as non-consecutive-entity.\newline
- Object: Artifacts or groups of artifacts clearly identified with a name, such as ships, weapons, statues, columns, dedications, etc.\newline
- Miscellaneous: Entities that do not (yet) have a specific first-level tag among those provided.\newline
- Time: Any absolute date or time expression.\newline
- Work: Titles of literary or non-literary works, in any form.\newline
\newline
\#\#\# Second Level Tags\newline
The following is the list of second level labels and their description:\newline
- .ancestry (collective.ancestry, person.ancestry): A designation or expression that refers unambiguously to one individual or group of individuals by using a family name, patronymic, matronymic, or other indication of lineage or familial relationships.\newline
- .animal (collective.animal, creature.animal): A type of creature or collective of creatures clearly identifiable with an animal or animal species.\newline
- .astronomy (creature.astronomy, collective.astronomy, place.astronomy): Named stars, groups of stars, constellations, and planets.\newline
- .author (person.author): A person clearly mentioned in relation to works they have authored. This tag may be modified or even omitted for project-specific goals.\newline
- .derivative (collective.derivative, person.derivative, place.derivative): An adjective derived from a toponym, personal name, or group name, used to identify things that are not individuals or collectives (for individuals or collectives, see .ethnic). Only the derivative is annotated, as the common noun in the expression does not act as a rigid designator. The first-level tag depends on the name from which the adjective derives (e.g. “Iberian” will be a place, “Platonic” a person, etc.).\newline
- .ethnic (collective.ethnic, person.ethnic): An ethnonym, demonym, or other word used to identify persons or collectives by means of their membership to a geographically or ethnically defined group. This tag is exclusively used with persons or collectives, as ethnics are mainly used in the ancient world to identify individuals via ethnic memberships (see also our rationale below). For all other uses of adjectives derived from places, use the .derivative subtag.\newline
- .epithet (collective.epithet, person.epithet, place.epithet): A capitalized epithet used to refer unambiguously to one individual, location, or collective, including nicknames, titles, and other appellatives.\newline
- .organization (collective.organization): Collectives identified by precise organizational structures, such as priesthoods, legions, religious, intellectual, or political groups and institutions, and so on.\newline
\newline
---\newline
\newline
\#\#\# Guidelines for Output:\newline
- For each word in the sentence, output the word followed by its IOB label and entity type where appropriate.\newline
- Use only the set of labels as defined earlier.\newline
- Each word should be on a new line in the same way it was input.\newline
- Do not add any other explanations or text.\newline
\newline
---\newline
\newline
\#\#\# Example:\newline
\newline
-> Example 1\newline
\#\#\#\#\# Input:\newline
\{INPUT\_EXAMPLE\_1\}\newline
\newline
\#\#\#\#\# Output:\newline
\{OUTPUT\_EXAMPLE\_1\}\newline
\newline
-> Example 2\newline
\#\#\#\#\# Input:\newline
\{INPUT\_EXAMPLE\_2\}\newline
\newline
\#\#\#\#\# Output:\newline
\{OUTPUT\_EXAMPLE\_2\}\newline
\newline
-> Example 3\newline
\#\#\#\#\# Input:\newline
\{INPUT\_EXAMPLE\_3\}\newline
\newline
\#\#\#\#\# Output:\newline
\{OUTPUT\_EXAMPLE\_3\}\newline
\newline
---\newline
\#\#\# Input:\newline
\{INPUT\_SENTENCE\}\newline
\newline
\#\#\# Output:\newline
} \\
\hline
\end{tabular}}
\caption{Few-Shot prompt template for fine-grained NER.}
\label{few-fine}
\end{table}

\end{document}